\documentclass{article}
\usepackage{nips12submit_e,times}
\usepackage{amsmath,amssymb}
\usepackage{multirow}
\usepackage{url}
\usepackage{booktabs}

\usepackage{mymath}
\usepackage{graphicx}

\graphicspath{{./}}

\begin{document}

\title{The Manifold of Human Emotions}

\author{
  Seungyeon Kim, Fuxin Li, Guy Lebanon, Irfan Essa\\
  College of Computing\\
  Georgia Institute of Technology\\
  \texttt{\{seungyeon.kim@, fli@cc., lebanon@cc., irfan@cc.\}gatech.edu}
}

\nipsfinalcopy

\maketitle

\begin{abstract}
  Sentiment analysis predicts the presence of positive or negative emotions in a text document. In this paper, we consider higher dimensional extensions of the sentiment concept, which represent a richer set of human emotions. Our approach goes beyond previous work in that our model contains a continuous manifold rather than a finite set of human emotions. We investigate the resulting model, compare it to psychological observations, and explore its predictive capabilities.
  \end{abstract}

\section{Introduction}

Sentiment analysis predicts the presence of a positive or negative emotion $y$ in a text document $x$. Despite its successes in industry, sentiment analysis is limited as it flattens the structure of human emotions into a single dimension. An alternative that has attracted a few researchers in recent years is to construct a finite collection of emotions and fit a predictive model for each emotion (see for example \cite{Mishne2005}). 
There are several significant difficulties with the above approach. First, it is hard to capture a complex statistical relationship between a large number of binary variables (representing emotions) and a high dimensional vector (representing the document). It is also hard to imagine a reliable procedure for compiling a finite list of all possible human emotions. Finally, it is not clear how to use documents expressing a certain emotion.Using labeled documents only in fitting models predicting their denoted labels ignores the relationship among emotions, and is problematic for emotions with only a few annotated.

We propose an alternative approach that models a stochastic relationship between the document $X$, an emotion label $Y$ (such as \texttt{sleepy} or \texttt{happy}), and a position on the mood manifold $Z$. We assume that all the emotional aspects in the documents are captured by the manifold, implying that the emotion label $Y$ can be inferred directly from the projection $Z$ of the document on the manifold, without needing to consult the document again.

\section{The Statistical Model}

We make the following four modeling assumptions concerning the document $X$, the discrete emotion label $Y\in\{1,2,\ldots,C\}$, and the position on the continuous mood manifold $Z\in\mathbb{R}^l$. 

\begin{enumerate}
\item We have the graphical structure: $X \to Z \to Y$, 
implying that the emotion label $Y\in \{1 ,\ldots,C\}$ is independent of the document $X$ given $Z$.
\item The distribution of $Z \in \mathbb{R}^l$ given a specific emotion label $Y=y$ is Gaussian:\\$\{Z|Y=y\} \sim \mathcal{N}(\mu_y,\Sigma_y)$
\item The distribution of $Z$ given the document $X$ (typically in a bag of words or $n$-gram representation) is a linear regression model: $\{Z|X=x\} \sim \mathcal{N}(\theta^{\top}x,\Sigma_x)$.
\item The distances between the vectors in $\{ \E(Z|Y=y): y\in C\}$ are similar to the corresponding distances in $\left\{\E(X|Y=y): y\in C\right\}$. \end{enumerate}
We make the following observations. First, the first assumption implies that the emotion label $Y$ is simply a discretization of the continuous $Z$. It is consistent with well known research in psychology (see Section 2) and with random projection theory, which state that it is often possible to approximate high dimensional data by projecting it on a low dimensional continuous space. Second, while $X$, $Y$ are high dimensional and discrete, $Z$ is low dimensional and continuous. This, together with the conditional independence in assumption 1 above, implies a higher degree of accuracy than modeling directly $X\to Y$. Intuitively, the number of parameters is on the order of $\text{dim}(X)+\text{dim}(Y)$ as opposed to $\text{dim}(X)\text{dim}(Y)$. Third, the Gaussian models in assumptions 2 and 3 are simple, and lead to efficient computational procedures. We also found them to work well in our experiments. The model may be easily adapted, however, to more complex models such as mixture of Gaussians or non-linear regression models (for example, we experimented with quadratic regression models). Fourth, assumption 4 suggests that we can estimate $\E(Z|Y=y)$ for all $y\in C$ via multidimensional scaling. MDS finds low dimensional coordinates for a set of points that approximates the spatial relationship between the points in the original high dimensional space. Lastly, the models in assumptions 2 and 3 are statistical and can be estimated from data using maximum likelihood.

\begin{figure*}
  \centering
  \includegraphics[width=.45\linewidth,trim=1.5em 0 1.5em 2.5em,clip]{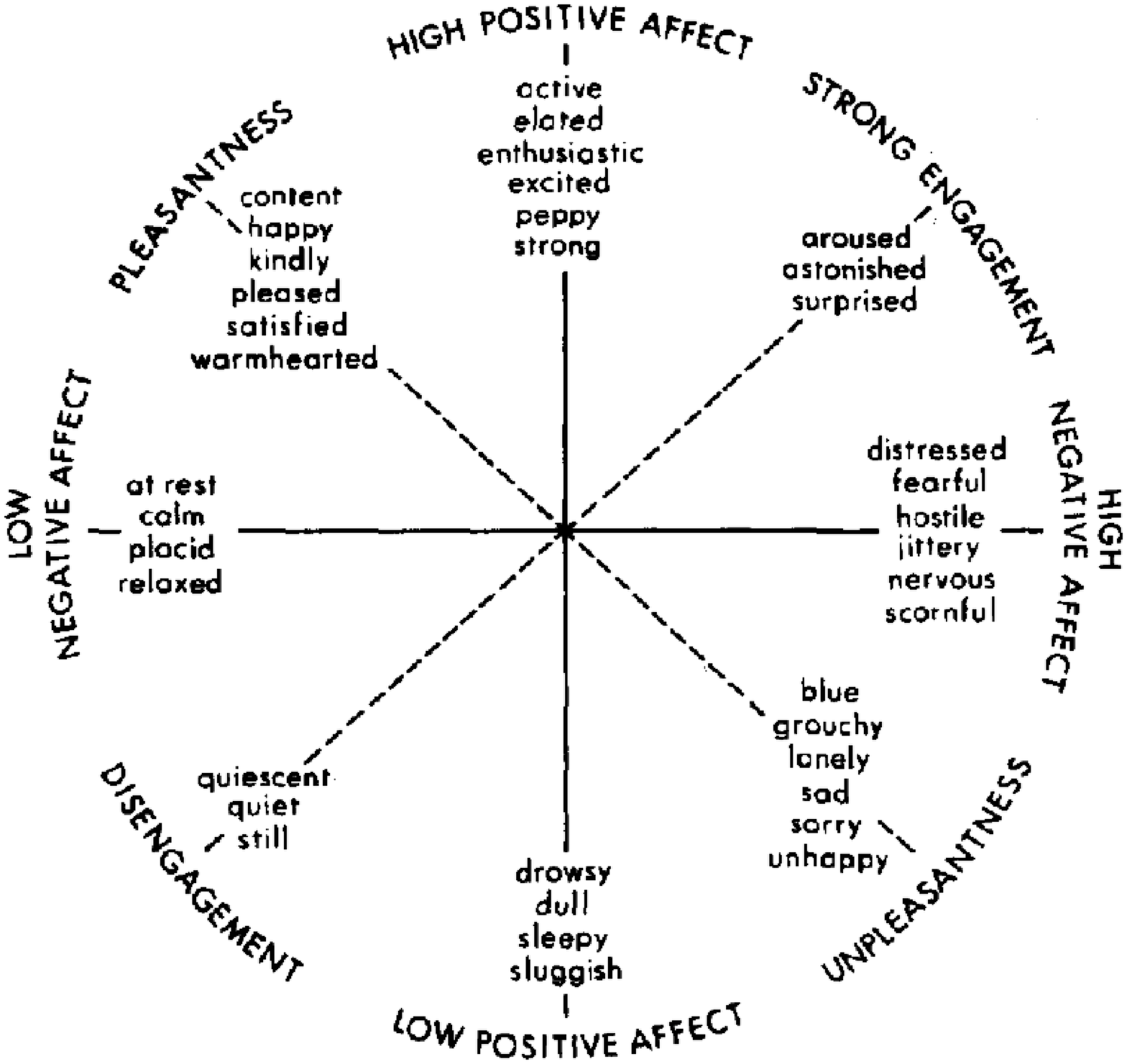}
  \includegraphics[width=.45\linewidth,trim=0.3em 0 0.6em 0,clip]{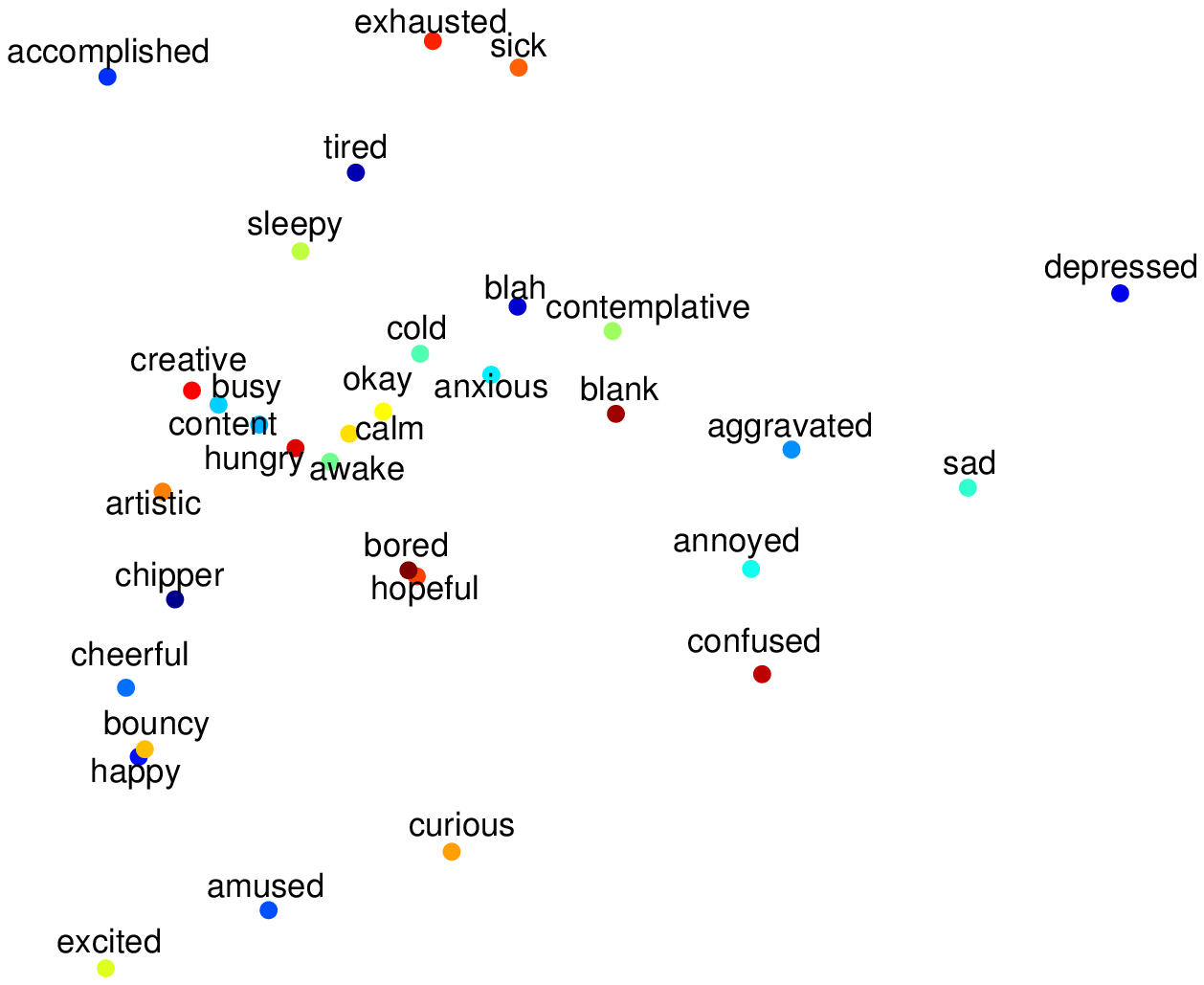}
  \vspace{-1em}
  \caption{\small (left) The two-dimensional structure of emotions from \cite{Watson1985}. We can interpret top-left to bottom-right axis as expressing sentiment polarity and the top-right to bottom-left axis as expressing engagement. (right) Mood centroids $\E(Z|Y=y)$ on the two most prominent dimensions in emotion space fitted from blog posts. See text for details.}\label{fig:2d_embedding}
      \vspace{-1.0em}
\end{figure*}

Motivated by the fourth modeling assumption, we determine the parameters $\mu_y=\E(Z|Y=y), y\in C$ by running multidimensional scaling (MDS) or Kernel PCA on the empirical versions of $\{\E(X|Y=y): y\in C\}$.

We estimate the parameter $\theta$, defining the regression $X\to Z$, by maximizing the likelihood which requires integrating over $Z\in\R^l$, a computationally difficult task when $l$ is not very low. We make use of approximating the Gaussian pdf with Dirac's delta function on $p(z)$,
\vspace{-1em}
\begin{align} 
  \hat\theta   &\approx \argmax_{\theta} \sum_i \log \frac{p(y^{(i)}) p_{\theta}({z^{(i)}}^* |x^{(i)})}{\sum_y p({z^{(i)}}^*|y) p(y)} 
  = \argmax_{\theta} \sum_i \log p_{\theta}({z^{(i)}}^* |x^{(i)})\vspace{-0.5em} \label{eq:dirac_delta_approx}\vspace{-1em}
\end{align}
where ${z^{(i)}}^* = \argmax_z p(z|y^{(i)}) = E(Z|y^{(i)})$,
which is equivalent to a least squares regression.

One interpretation of our model $X\to Z\to Y$ is that $Z$ forms a sufficient statistic of $X$ for $Y$. We can thus consider adapting a wide variety of predictive models (for example, logistic regression or SVM) on $Z\mapsto Y$. These discriminative classifiers are trained on $\{(\hat{Z}^{(i)},Y^{(i)}), i=1,\ldots,n\}$.

\section{Experiments}

We used crawled Livejournal\footnote{\url{http://www.livejournal.com}} data as the main dataset. About 20\% of the blog posts feature these optional annotations in the form of emoticons. The annotations may be chosen from a pre-defined list of possible emotions, or a novel emotion specified by the author. We crawled 15,910,060 documents and selected 1,346,937 documents featuring the most popular 32 emotion labels (in respect to the number of documents annotated in).

In Figure~\ref{fig:2d_embedding}, we compare our model to Watson and Tellegen's well known psychological model. We make the following observations. First, the horizontal axis expresses a sentiment polarity-like emotion. The left part features emotions such as \texttt{accomplished}, \texttt{happy} and \texttt{excited}, while the right part features emotions such as \texttt{sad} and \texttt{depressed}. This is in agreement with Watson and Tellegen's observations. Second, the vertical axis expresses the level of mental engagement or energy level. The top part features emotions such as \texttt{exhausted} or \texttt{tired}, while the bottom part features emotions such as \texttt{curious} or \texttt{excited}. This agrees partially with the engagement dimension in the psychological model. Third, the neutral moods \texttt{blank}, stay in the middle of the picture. This agreement between our mood manifold and the psychological findings is remarkable in light of the fact that the two models used completely different experimental methodology (blog data vs. surveys).

\begin{table*}
  \caption{\small Macro F1 score and accuracy over the test set in multiclass emotion classification over top 32 moods (left) and sentiment polarity task (right): \{cheerful, happy, amused\} vs \{sad, annoyed, depressed, confused\}. See text for details.} \label{tbl:classification_mood}
  \scriptsize
  \begin{tabular*}{.48\linewidth}{@{\extracolsep{\fill}} l l r r r r}
    \toprule
    &  & \multicolumn{2}{c}{Original Space} &  \multicolumn{2}{c}{Mood Manifold}\\
    &  &  F1 & Acc. & F1 & Acc. \\
    \midrule
    LDA & full   &    n/a &    n/a & \bf 0.1247 & \bf 0.1635 \\
        & diag.  & \bf 0.1229 & 0.1441 & 0.1160 & \bf 0.1600 \\
        & spher. & 0.0838 & 0.1075 & \bf 0.0896 & \bf 0.1303 \\
    QDA & full   &    n/a &    n/a & \bf 0.1206 & \bf 0.1478 \\
        & diag.  & 0.0878 & 0.0931 & \bf 0.1118 & \bf 0.1463 \\
        & spher. & 0.0777 & 0.0989 & \bf 0.0873 & \bf 0.1253 \\
    Log.Reg. &   & 0.1231 & 0.1360 & \bf 0.1477 & \bf 0.1667 \\
    \bottomrule    
  \end{tabular*}
  \quad
  \begin{tabular*}{.48\linewidth}{@{\extracolsep{\fill}} l l r r r r}
    \toprule
    &  & \multicolumn{2}{c}{Original Space} &  \multicolumn{2}{c}{Mood Manifold}\\
    &  &  F1 & Acc. & F1 & Acc. \\
    \midrule
    LDA & full   &    n/a &    n/a & \bf 0.7340 & \bf 0.7812 \\
        & diag.  & 0.7183 & 0.7436 & \bf 0.7365 & \bf 0.7663 \\
        & spher. & 0.6358 & 0.6553 & \bf 0.7482 & \bf 0.7699 \\
    QDA & full   &    n/a &    n/a & \bf 0.6500 & \bf 0.7446 \\
        & diag.  & 0.6390 & 0.6398 & \bf 0.6704 & \bf 0.7510 \\
        & spher. & 0.6091 & 0.6143 & \bf 0.7472 & \bf 0.7734 \\
    Log.Reg. &   & 0.7350 & 0.7624 & \bf 0.7509 & \bf 0.7857\\
    \bottomrule
  \end{tabular*}
\end{table*}

We performed emotion classification experiment (Table~\ref{tbl:classification_mood}, left) on the Livejournal data. We considered the goal of predicting the most popular 32 moods. We also considered a binary classification tasks obtained by partitioning the set of moods into two clusters.

Table~\ref{tbl:classification_mood} compare classification results using the original bag of words feature space and the manifold model, using different types of classification methods: LDA, QDA with different covariance matrix models, and logistic regression. Bold faces are improvements over the baseline with statistical significance of $t$-test of random trials. Most of experimental results show that the mood manifold model results in statistically significant improvements than using original bag of words feature.

\section{Summary}

In this paper, we introduced a continuous representation for human emotions $Z$ and constructed a statistical model connecting it to documents $X$ and to a discrete set of emotions $Y$. Our fitted model bears close similarities to models developed in the psychological literature, based on human survey data. Several attempts were recently made at inferring insights from social media or news data through sentiment prediction. It is likely that the current multivariate view of emotions will help make progress on these important and challenging tasks.

\small
\bibliographystyle{plain}
\bibliography{../../share/externalPapers,../../share/groupPapers}

\end{document}